\documentclass[10pt,a4paper]{article}
\usepackage[a4paper,left=25mm,right=25mm,top=21mm,bottom=24mm,footskip=10mm]{geometry}
\usepackage{fontspec}
\newfontfamily\marcellus{MarcellusSC-Regular.ttf}[Path=fonts/]
\usepackage{amsmath}
\usepackage{microtype}
\usepackage{graphicx}
\usepackage{xcolor}
\usepackage{booktabs,tabularx,array}
\usepackage[font=small,labelfont=bf,labelsep=colon,skip=5pt]{caption}
\usepackage{wrapfig}
\usepackage{enumitem}
\usepackage{titlesec}
\usepackage[most]{tcolorbox}
\usepackage[skip=5.5pt plus 1.5pt minus 0.5pt,indent=0pt]{parskip}
\usepackage[colorlinks,allcolors=navy,pdftitle={FDE-Gym: The Last Mile of AI Delivery}]{hyperref}

\definecolor{navy}{HTML}{1D4F91}
\definecolor{hair}{HTML}{C9CED6}
\titleformat{\section}{\fontsize{13}{15}\bfseries}{\thesection}{0.75em}{}
\titlespacing*{\section}{0pt}{14pt plus 3pt minus 2pt}{5pt plus 1pt}
\titleformat{\subsection}{\fontsize{10.6}{12.5}\bfseries}{\thesubsection}{0.7em}{}
\titlespacing*{\subsection}{0pt}{9pt plus 2pt minus 1pt}{2.5pt plus 1pt}
\setlist[itemize]{leftmargin=5.2mm,itemsep=1.5pt,topsep=2pt,parsep=0pt,label={\color{navy}\textbullet}}

\newtcolorbox{finding}[1]{enhanced,frame hidden,interior hidden,boxrule=0pt,arc=0pt,left=4.2mm,right=0pt,top=0.5mm,bottom=0.6mm,
  borderline west={1.15mm}{0pt}{navy},before skip=7pt plus 1pt,after skip=7pt plus 1pt,
  fontupper=\fontsize{10.2}{12.4}\selectfont\raggedright,before upper={{\marcellus\color{navy}\fontsize{9.4}{12}\selectfont #1}\hspace{2mm}}}

\newenvironment{paperabstract}{\begin{center}\textbf{\fontsize{11}{13}\selectfont Abstract}\end{center}\vspace{1pt}%
  \begin{list}{}{\leftmargin=7mm\rightmargin=7mm\topsep=0pt\parsep=4.5pt\listparindent=0pt}\item\relax\fontsize{9.6}{11.3}\selectfont}{\end{list}\vspace{2pt}}
\newenvironment{references}{\begin{enumerate}[label={[\arabic*]},leftmargin=7mm,labelsep=1.6mm,itemsep=1.2pt,topsep=2pt,parsep=0pt]\fontsize{8.6}{10.4}\selectfont\raggedright}{\end{enumerate}}
\newcommand{\linksep}{\hspace{2.2mm}{\color{hair}|}\hspace{2.2mm}}

\newsavebox{\headbox}\newsavebox{\bleedbox}\newlength{\bleedshift}
\newcommand{\frontpage}[7]{%
  \thispagestyle{plain}%
  \sbox{\bleedbox}{\includegraphics[width=\dimexpr\textwidth+#5\textwidth+#5\textwidth\relax]{#4}}%
  \setbox\headbox=\vbox{\hsize=\textwidth\parskip=0pt\centering
    {\marcellus\fontsize{19}{23}\selectfont #1\par}\vspace{2mm}
    {\fontsize{8}{10}\selectfont\sffamily #2\par}\vspace{3mm}
    #3}%
  \setlength{\bleedshift}{\dimexpr\ht\headbox+\dp\headbox+2.5mm-#6\textwidth+\ht\bleedbox\relax}%
  {\offinterlineskip
   \hbox to \textwidth{\smash{\rlap{\hspace{-#5\textwidth}\raisebox{-\bleedshift}{\usebox{\bleedbox}}}}\hfil}%
   \box\headbox}%
  \par\vspace{\dimexpr 2.5mm+\ht\bleedbox-#6\textwidth-#6\textwidth\relax}%
  \noindent\begin{minipage}{\textwidth}\captionof{figure}{#7}\end{minipage}\par
  \frontsecond}
\newcommand{\frontsecond}[3]{\vspace{1.5mm}\noindent\begin{minipage}{\textwidth}\centering\includegraphics[width=#2\textwidth]{#1}\captionof{figure}{#3}\end{minipage}\clearpage}

\begin{document}
\frontpage{FDE-Gym: The Last Mile of AI Delivery}{\href{https://fde-gym.com/paper}{Website}\linksep \href{https://fde-gym.com}{Online Gym}\linksep \href{https://fde-gym.com/paper/\#leaderboard}{Leaderboard}\linksep \href{https://github.com/deeplethe/fde-gym}{GitHub}\linksep \href{https://huggingface.co/datasets/DeepLethe/FDE-Gym}{Hugging Face}\linksep \href{https://fde-gym.com/paper/\#submit}{Submission}\linksep \href{mailto:info@deeplethe.com}{Email us}}{%
{\fontsize{9.6}{12}\selectfont\bfseries Core Authors\par}\vspace{0.5mm}
{\fontsize{10.4}{14.5}\selectfont \mbox{Dongxu Yang\textsuperscript{1,\ensuremath{\ast}}}, \mbox{Hongda Yan\textsuperscript{1}}, \mbox{Renxin Wang\textsuperscript{1}}, \mbox{Baoli Zhang\textsuperscript{8}}, \mbox{Haojia Zuo\textsuperscript{2}}, \mbox{Ying Zhao\textsuperscript{2}}, \mbox{Youmin Chen\textsuperscript{3}}\par}\vspace{0.2mm}
{\fontsize{7.8}{10}\selectfont \textsuperscript{\ensuremath{\ast}}Corresponding author.\par}\vspace{2.4mm}
{\fontsize{9.6}{12}\selectfont\bfseries Contributors\par}\vspace{0.5mm}
{\fontsize{9}{13.2}\selectfont \mbox{Xiaojian Liao\textsuperscript{4}}, \mbox{Congming Gao\textsuperscript{5}}, \mbox{Tianyang Jiang\textsuperscript{15}}, \mbox{Yixiao Chen\textsuperscript{11}}, \mbox{Qing Wang\textsuperscript{6}}, \mbox{Zhiguo Yu\textsuperscript{16}}, \mbox{Xiaqing Li\textsuperscript{7}}, \mbox{Yuhao Zhang\textsuperscript{9}}, \mbox{Jiawen Sun\textsuperscript{17}}, \mbox{Renzhe Xu\textsuperscript{10}}, \mbox{De Mi\textsuperscript{19}}, \mbox{Tiancheng Wu\textsuperscript{18}}, \mbox{Xianya Fu\textsuperscript{12}}, \mbox{Xuanlu Zhao\textsuperscript{13,20}}, \mbox{Bo Feng\textsuperscript{21}}, \mbox{Meixuan Ren\textsuperscript{22}}, \mbox{Guanming Lin\textsuperscript{14}}, \mbox{Ziheng Zhou\textsuperscript{23}}, \mbox{Xinhua Wang\textsuperscript{23}}\par}\vspace{2mm}
{\fontsize{6.8}{10}\selectfont \mbox{\textsuperscript{1}DeepLethe}\hspace{0.95em plus 0.5em} \mbox{\textsuperscript{2}Tsinghua University}\hspace{0.95em plus 0.5em} \mbox{\textsuperscript{3}Shanghai Jiao Tong University}\hspace{0.95em plus 0.5em} \mbox{\textsuperscript{4}Beihang University}\hspace{0.95em plus 0.5em} \mbox{\textsuperscript{5}Xiamen University}\hspace{0.95em plus 0.5em} \mbox{\textsuperscript{6}Nanjing University}\hspace{0.95em plus 0.5em} \mbox{\textsuperscript{7}Beijing Jiaotong University}\hspace{0.95em plus 0.5em} \mbox{\textsuperscript{8}Beijing University of Posts and Telecommunications}\hspace{0.95em plus 0.5em} \mbox{\textsuperscript{9}Tianjin University}\hspace{0.95em plus 0.5em} \mbox{\textsuperscript{10}Shanghai University of Finance and Economics}\hspace{0.95em plus 0.5em} \mbox{\textsuperscript{11}Harvard University}\hspace{0.95em plus 0.5em} \mbox{\textsuperscript{12}Stanford University}\hspace{0.95em plus 0.5em} \mbox{\textsuperscript{13}University College London}\hspace{0.95em plus 0.5em} \mbox{\textsuperscript{14}Cornell University}\hspace{0.95em plus 0.5em} \mbox{\textsuperscript{15}Huawei}\hspace{0.95em plus 0.5em} \mbox{\textsuperscript{16}Atlassian}\hspace{0.95em plus 0.5em} \mbox{\textsuperscript{17}Meta}\hspace{0.95em plus 0.5em} \mbox{\textsuperscript{18}Microsoft Research Asia}\hspace{0.95em plus 0.5em} \mbox{\textsuperscript{19}StellarCube.ai}\hspace{0.95em plus 0.5em} \mbox{\textsuperscript{20}UWits Technology (Beijing) Co., Ltd.}\hspace{0.95em plus 0.5em} \mbox{\textsuperscript{21}Bopai Management Consulting (Shenzhen) Co., Ltd.}\hspace{0.95em plus 0.5em} \mbox{\textsuperscript{22}Exploration and Development Research Institute, PetroChina Southwest Oil \& Gasfield Company}\hspace{0.95em plus 0.5em} \mbox{\textsuperscript{23}Kunlun Digital Technology Co., Ltd.}\par}}
{fig1}{0.07}{0.05}{Construction of FDE-Gym. Experts adapt 20 real-world delivery projects from 18 sectors into operational customer environments, alter key facts to create variant tasks, and retain the 95 tasks whose graders have been validated against the predefined scoring anchors.}%
{fig2}{0.860}{Model performance across task instruction levels. \textbf{(a)} Mean L3 score for each model on all 95 tasks, on a normalized scale where 0 represents the do-nothing baseline and 1 represents the reference solution. Solid bars show scores after deducting stakeholder contact cost; lighter bars show normalized business uplift before this deduction. \textbf{(b)} Scores as progressively less information is provided in the task instruction. L0-L2 are evaluated on the 20 original tasks and L3 on all 95 tasks.}

\begin{paperabstract}
Existing agent benchmarks typically evaluate agents on well-specified tasks defined by issue reports, specifications, or task tickets. Real-world customer delivery, however, poses fundamentally different challenges: forward-deployed engineers (FDEs) must clarify ambiguous requirements, work in live systems, and uncover unstated constraints. To address this gap, we introduce FDE-Gym, a benchmark of 95 end-to-end delivery tasks from 20 customer scenarios across 18 sectors, each adapted by industry experts from real-world delivery projects. In simulated customer environments featuring scripted stakeholders and modifiable customer systems, agents interact with stakeholders and operate on production-like systems to deliver working solutions. Beyond binary task completion, FDE-Gym scores agents on how much their delivered system improves the customer’s own success metric, normalized between 0 (unchanged system) and 1 (reference solution); scores are penalized for stakeholder interactions (messages, trial runs, and waiting time) to discourage excessive or careless engagement. FDE-Gym also provides an online platform where humans can develop and practice FDE skills under the same grading criteria.

We evaluate 13 model configurations from seven providers. In the realistic setting (unstated constraints need to be discovered), the best configuration achieves a mean FDE-Gym score of 0.56, and resolves only 14 of the 95 tasks (a score of at least 0.8 with no incident). Discovering unstated constraints is the primary bottleneck: performance drops sharply when models must uncover them without explicit guidance. However, information discovery alone does not account for the gap: even on the 23 tasks where the best configuration obtains every key fact from stakeholders, it achieves a normalized business uplift of 0.8 or above on only 10.

These results reveal a substantial gap between solving well-specified tasks and delivering effective AI systems for real customers. Closing this gap requires agents that can surface unstated constraints, operate reliably in live systems, and achieve measurable improvements on customer metrics.
\end{paperabstract}

\section{Introduction}

Deploying AI in enterprise settings requires more than meeting stated customer requirements. The challenge often lies in identifying which changes will improve customer operations in practice. One scenario in our benchmark illustrates this problem. A vice president at a drilling operator seeks a generative-AI success story for the November board meeting. The vice president requests a daily summary email and an “Ask your wells” chat box, noting that the downtime-coding system is already in place and therefore outside the project scope.

Yet a solution built strictly to the task instruction can be delivered on time, pass every functional check, and still fail to move the customer’s own business metric by a single point. The bottleneck lies in monthly downtime credit claims. The existing keyword-based script incorrectly attributes some downtime to the contractor, resulting in at least one disputed line each month and suspending the credit for that month. Addressing this problem requires revisiting the downtime-coding system that the sponsor had explicitly excluded from the project scope. The engineer must identify the applicable contractual rule, confirm the interpretation with the cost engineer, and obtain approval for the change from an approver who is unavailable that week.

This is the role of a forward-deployed engineer (FDE), an engineer assigned by a vendor to work within a customer organization. The role involves more than building what the sponsor initially requests. %
An FDE must identify what will actually help and work with customer systems and stakeholders to deliver the improvement.

Existing benchmarks rarely address this last mile of AI delivery, typically assuming that the problem has already been well defined. Software-engineering benchmarks provide issues and hidden tests {\color{navy}[\hyperlink{ref.1}{1}, \hyperlink{ref.2}{2}, \hyperlink{ref.3}{3}]}, deployment benchmarks provide specifications {\color{navy}[\hyperlink{ref.4}{4}]}, enterprise benchmarks provide predefined tasks {\color{navy}[\hyperlink{ref.5}{5}, \hyperlink{ref.6}{6}]}, and clarification benchmarks withhold information from an otherwise correctly specified task {\color{navy}[\hyperlink{ref.7}{7}, \hyperlink{ref.8}{8}]} --- in each case, the underlying problem remains fixed.
Recent benchmarks come closer to real-world delivery.
\ensuremath{\tau}\textsuperscript{\ensuremath{\tau}}-bench {\color{navy}[\hyperlink{ref.9}{9}]} evaluates agents that build customer-service systems from client requirements and deploy the resulting systems against held-out users. A post-training benchmark {\color{navy}[\hyperlink{ref.10}{10}]} assigns agents the role of forward-deployed engineers and evaluates whether the delivered model improves performance. XYEval {\color{navy}[\hyperlink{ref.11}{11}]} tests agents on misleading user requests. However, these benchmarks still address individual aspects of delivery in isolation.

FDE-Gym addresses these challenges in a single benchmark of 95 tasks from 20 customer scenarios across 18 sectors, each adapted by industry experts from real-world delivery projects. The sponsor’s request may target the wrong problem, critical constraints must be discovered through stakeholder interaction, and performance is measured using the customer’s own business metric. A solution that addresses the wrong problem therefore earns no credit.

Delivered systems are deployed without modification and evaluated on held-out production data. The FDE-Gym score measures improvement on the customer’s own business metric after deducting stakeholder contact cost (Section 2.3 details what is charged).
The do-nothing baseline (the unchanged system) scores 0, the expert-authored reference solution scores 1, and literal compliance with the sponsor’s request scores at most 0.2. Experimental results show that current models perform well below the reference solution on this scale. %

To identify where agents fail, we present each scenario at four task instruction levels with progressively less information (from a fully specified problem (L0) down to the sponsor's original request, which may not address the underlying problem (L3)).
The environment, stakeholders, and grader remain unchanged across all four levels. Because only the task instruction varies, differences in performance isolate the effect of the information withheld. We evaluate 13 model configurations from seven providers using this setup. Five main findings emerge.

\begin{itemize}
\item \textbf{Even the best model configuration falls well short of the reference solution.} At L3, Claude Opus 5.5 achieves the highest mean score of 0.56 and resolves 14 of 95 tasks, while five of the thirteen configurations resolve none. Selecting the best-performing configuration for each task still yields a score of only 0.70, with 30 tasks resolved.
\item \textbf{Discovering unstated constraints is the primary bottleneck for leading models.} When the relevant rules and constraints are provided, the three leading models score 0.73–0.82. Requiring these models to discover the constraints reduces scores sharply to 0.39–0.62. On the original tasks, L3 scores are at most 0.08 points lower than L1 scores.
\item \textbf{Obtaining the necessary information does not ensure effective delivery.} At L3, leading models obtain 72–85\% of the key facts available from stakeholders, yet the delivered systems achieve only 0.60–0.70 of the reference improvement. Stakeholder contact cost reduces the final scores by a further 0.14–0.22.
\item \textbf{Weak models can leave customers worse off than if no changes had been made.} Claude Haiku 4.5 scores below the do-nothing baseline on 79 of 95 tasks, while systems delivered by Mistral Small 4 fall below this baseline on 51 tasks even before deducting stakeholder contact cost.
\item \textbf{No model performs well across all scenarios.} In five of the 20 scenarios, no model achieves a mean score of at least 0.5. Across tasks within a scenario, the median score range for a leading model is 0.34–0.44.
\end{itemize}

We make three contributions. First, we introduce FDE-Gym, a benchmark that evaluates whether agents can identify the underlying customer problem and deliver a working solution that measurably improves business outcomes. Second, we develop a methodology for constructing scenarios that combines task variants, scripted stakeholders, and predefined scoring anchors to support reproducible evaluation and limit rewards for literal compliance and shortcut solutions. Third, we evaluate 13 model configurations and provide an online platform where humans and agents can practice delivery under the same grading criteria.

\section{Anatomy of a Task}
Each task represents a single customer engagement involving one customer (the organization that hires the agent, whose people are the stakeholders the agent can question), one sponsor (the customer-side person who assigns the task), and one production system to improve. Figure 3 shows the four components of a task.

\begin{figure}[h]
\centering
\includegraphics[width=1.000\linewidth]{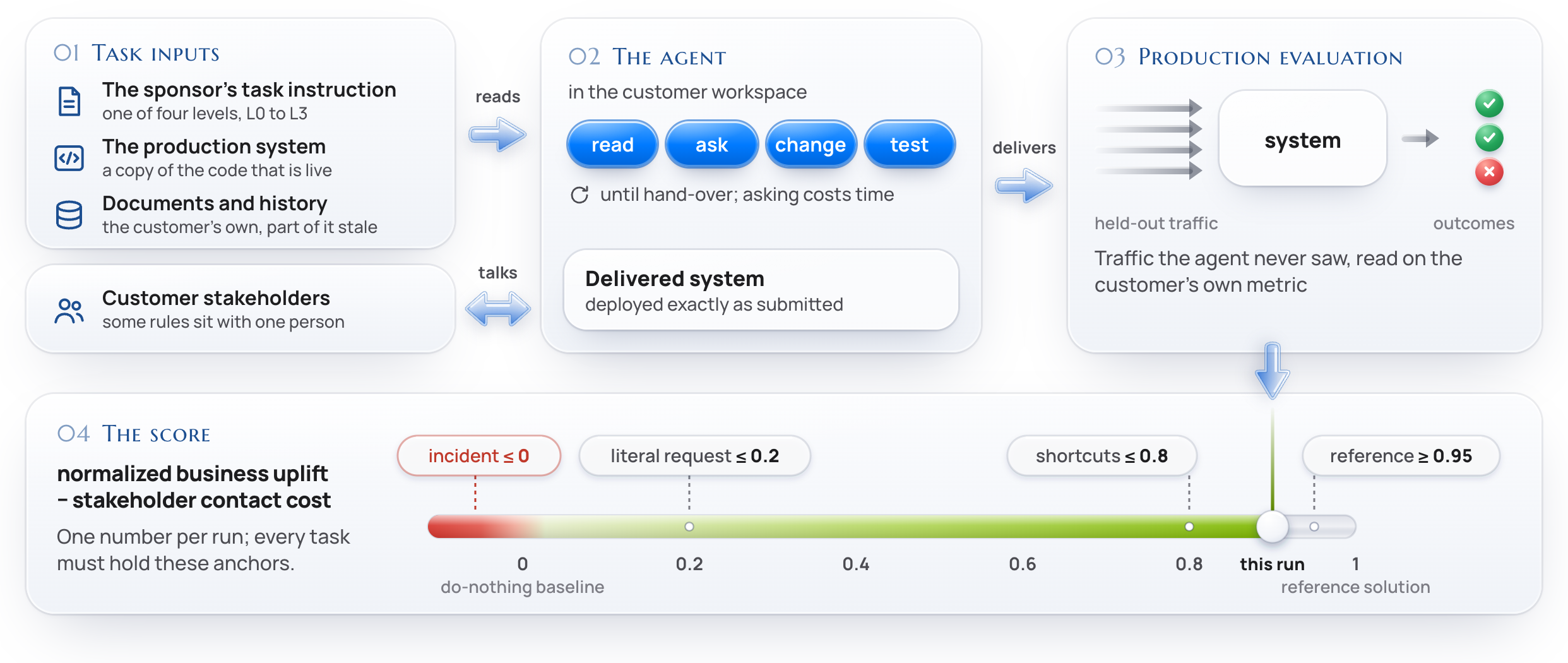}
\caption{Components of an FDE-Gym task, showing the inputs, agent actions and deliverable, deployment and evaluation of the delivered system, and interpretation of the resulting score. The agent reads the task inputs and exchanges messages with customer stakeholders, who answer questions and may also initiate contact. A task is included only if the scoring anchors shown on the axis are satisfied.}
\end{figure}

\subsection{Environment and Available Information}

The agent starts in a workspace containing the task instruction (the description of the work assigned by the sponsor), a copy of the customer’s production system, relevant documents and historical data, and a staff directory with contact details. The runtime environment provides standard Python and SQLite. Installing additional packages and accessing the general network are prohibited. The delivered system may send network requests only to the company’s LLM gateway, which charges according to usage. Once delivery is complete, the contents of \texttt{system/} are deployed to production without modification. Production continues to call the original entry points, so new functionality must be integrated with these entry points to take effect.

The agent is not explicitly informed that stakeholder contact incurs costs, that some people should not be contacted or are irrelevant to the problem, or that the sponsor’s request may be misdirected. Identifying these issues is a central objective of the evaluation. Relevant clues appear in the staff directory, customer documents, and data, where an FDE would typically look for such information. As in real-world projects, documents may be outdated, legacy code may reflect incorrect assumptions, and important information may not be explicitly stated in the available materials.

\subsection{Task Instruction Levels}

\begin{table}[tbp]
\caption{The four task instruction levels. The workspace, stakeholders, and grading criteria remain identical across levels; only the task instruction varies.}
\centering\small
\begin{tabularx}{\linewidth}{@{}l>{\raggedright\arraybackslash}X>{\raggedright\arraybackslash}X@{}}
\toprule
\textbf{Level} & \textbf{Information provided in the task instruction} & \textbf{What the agent must determine} \\
\midrule
\textbf{L0} & The underlying problem, applicable rules and constraints, and how the result is scored & How to design and implement an effective solution \\
\textbf{L1} & A clear statement of the underlying problem & The applicable rules and constraints, and how to design and implement a solution \\
\textbf{L2} & A vague but directionally correct request & The problem to be solved, the applicable constraints, and how to design and implement a solution \\
\textbf{L3} & The sponsor’s original request, which may target the wrong problem & Whether the request addresses the underlying problem, which constraints apply, and how to design and implement a solution \\
\bottomrule
\end{tabularx}
\end{table}

The four task instruction levels (Table 1) distinguish the challenges of implementation, constraint discovery, and problem clarification by varying the task instruction while keeping the environment and evaluation criteria fixed. The L0 instruction explicitly states the underlying problem, relevant rules and constraints, and scoring criteria, leaving the agent to design and implement a solution. The L1 instruction withholds the rules and constraints, while the L2 instruction also provides only a vague description of the problem. The L3 instruction presents the sponsor’s original request, requiring the agent to assess whether the request addresses the underlying problem. We use L3 as the primary evaluation setting because this level most closely reflects the starting point of real-world customer delivery projects. L0 does not eliminate the need for engineering decisions. By specifying the rules while leaving solution design open, L0 measures what an agent can build when all underlying facts are provided.

\subsection{FDE-Gym Score}

FDE-Gym evaluates a delivered system by the improvement it achieves on the customer’s business metric, measured on held-out production data. For example, an after-sales system is assessed by its net financial benefit, while a drilling system is assessed by the downtime hours correctly attributed to a contractor for compensation claims.

We define the following metrics to assess agent performance on these delivery tasks.

\textbf{Normalized Business Uplift (\textit{U}).} We measure the delivery’s business improvement relative to that achieved by an expert-authored reference solution:

\begin{equation}
\label{eq:business-uplift}
U = \frac{\mathrm{KPI}_{\mathrm{delivered}}-\mathrm{KPI}_{\mathrm{baseline}}}{\mathrm{KPI}_{\mathrm{reference}}-\mathrm{KPI}_{\mathrm{baseline}}}
\end{equation}

Here, KPI\textsubscript{baseline}, KPI\textsubscript{delivered}, and KPI\textsubscript{reference} denote the customer’s business metric for the unchanged system, the delivered system, and the expert-authored reference solution, respectively, evaluated on the same held-out data. The unchanged system is the do-nothing baseline. Thus, \textit{U} = 0 represents no improvement, \textit{U} = 1 matches the reference improvement, and \textit{U} \ensuremath{<} 0 indicates an outcome worse than the baseline.

\textbf{Incident Adjustment (\textit{U}\textsubscript{adj}).} If a predefined operational incident occurs, such as an unauthorized action, positive uplift is set to zero while negative uplift is retained:

\begin{equation}
\label{eq:incident-adjustment}
U_{\mathrm{adj}} = \begin{cases} U, & \text{if no incident occurs}, \\ \min(U,0), & \text{otherwise}. \end{cases}
\end{equation}

\textbf{Stakeholder Contact Cost (\textit{C}).} Stakeholder contact cost is expressed in the same normalized units as business uplift. It is the sum of the charges incurred by the agent’s contacts with the customer:

\begin{equation}
\label{eq:contact-cost}
C = \sum_{k\in K} c_k,
\end{equation}

where \textit{K} is the set of chargeable contacts in the run and \textit{c}\textsubscript{k} is the charge for contact \textit{k} under the task’s predefined pricing and contact rules. Each stakeholder has a hidden per-message rate, with higher charges when the agent exceeds their patience or violates a contact restriction. Unanswered messages are not charged. In scenarios with a trial environment or a calendar, each trial run and each day of waiting is also a chargeable contact.

\textbf{FDE-Gym Score (\textit{S}).} The FDE-Gym Score measures delivery performance in a single evaluation run. It is the incident-adjusted normalized business uplift minus stakeholder contact cost:

\begin{equation}
\label{eq:final-score}
S = U_{\mathrm{adj}}-C.
\end{equation}

Without an incident, \textit{S} = \textit{U} \ensuremath{-} \textit{C}. A task is considered \textit{resolved} if its score is at least 0.8 and no incident occurs.

\textbf{Average FDE-Gym Score (\ensuremath{\overline{S}}).} The Average FDE-Gym Score summarizes a model’s delivery performance across tasks and repeated runs. To prevent extreme business losses from dominating the reported average, we floor the incident-adjusted uplift at \ensuremath{-}1 before deducting the full stakeholder contact cost. We then average across tasks and runs:

\begin{equation}
\label{eq:aggregation}
\overline{S} = \frac{1}{N}\sum_{i=1}^{N}\frac{1}{R}\sum_{r=1}^{R}\left[\max\left(U_{\mathrm{adj},i,r},-1\right)-C_{i,r}\right].
\end{equation}

Here, \textit{N} is the number of evaluated tasks, \textit{R} is the number of runs per task, and \textit{i} and \textit{r} index tasks and runs, respectively. Each task receives equal weight. The floor applies only to aggregation and does not change the single-run score \textit{S} used to determine whether a task is resolved.

Figure 4 presents the drilling scenario, its customer metric, and delivery results.

\begin{figure}[tbp]
\centering
\includegraphics[width=1.000\linewidth]{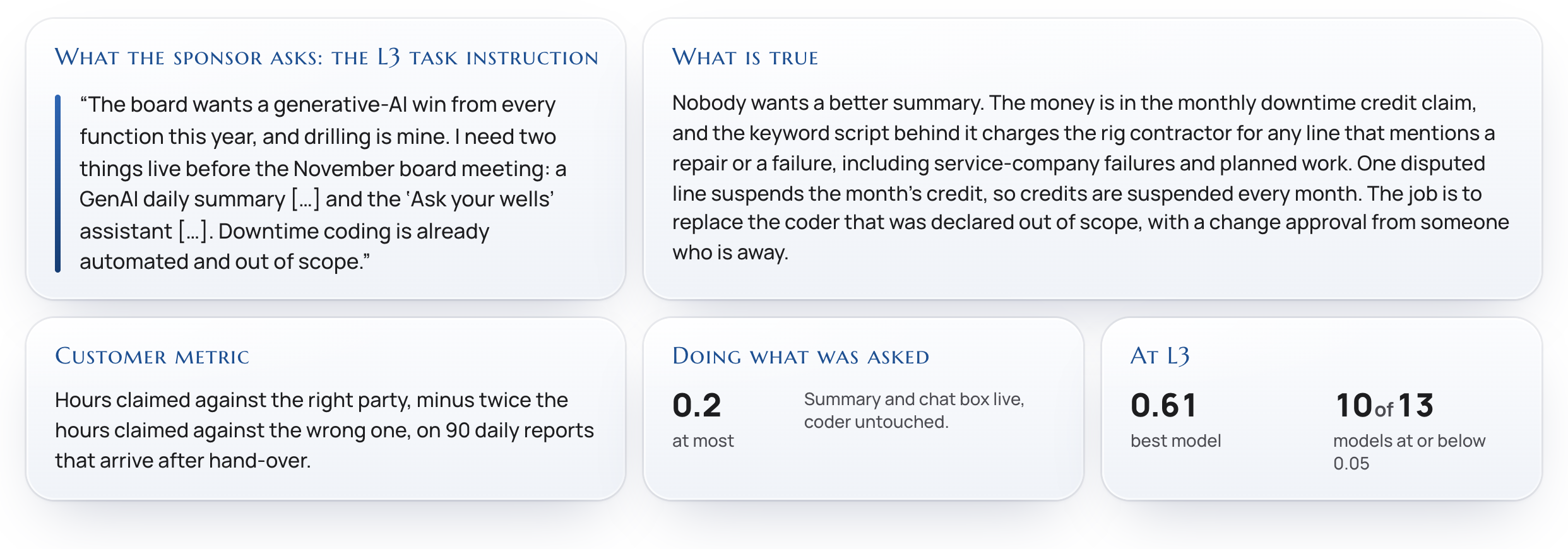}
\caption{A worked example: the drilling scenario (\texttt{oil\_npt}), original task.}
\end{figure}

\section{Building the Benchmark}

\subsection{Scenario Construction from Real-World Delivery Projects}

Figure 1 traces how a real delivery becomes a task in the benchmark. Industry experts adapted real-world customer delivery projects carried out in China, Singapore, and the United States since 2024 into 20 scenarios spanning 18 sectors (Table 2). Each scenario yields an original task and several variants of it, which share a common customer environment but differ in key facts. Each original task and each variant is a separate benchmark task, for a total of 95 tasks. Tasks require agents to deliver an agent, workflow, or service that operates in the customer’s environment. The scenarios are grounded in industry practice, with the relevant regulations, standards, and public sources documented. Companies and individuals are either de-identified or fictional, and some data are adapted from real-world industry sources.

\subsection{Scenario Variants and Underlying Facts}

A single fixed task per scenario may allow models to rely on memorized solutions or recurring patterns. To reduce this risk, experts designed two to seven variants in addition to each scenario’s original task. Variants change key facts, including which rule is outdated, who holds critical information, and whether the sponsor’s request addresses the underlying problem. These changes may require different delivery solutions, so a solution that succeeds on one task of a scenario may fail on another. Across scenarios, no single problem pattern appears in more than one-third of the set. Together, these design choices encourage agents to evaluate the evidence in each task rather than apply a fixed solution.

\subsection{Stakeholder Simulation}

FDE-Gym simulates more than 130 customer stakeholders across its 20 scenarios. Stakeholder responses are scripted for reproducibility, as switching the user-simulation model can shift measured agent success rates by several percentage points {\color{navy}[\hyperlink{ref.14}{14}]}. A small model maps questions to topics, while each stakeholder’s knowledge, misconceptions, and conditions for disclosing information are fixed in advance. The scripts incorporate limited patience, requirements for supporting evidence, dependencies on prior conversations, temporary unavailability, and contact restrictions. Agents must infer stakeholder roles and communication constraints from customer documents.

\begin{table}[tbp]
\caption{The 20 scenarios of FDE-Gym v1.0. A scenario’s tasks (the original and its variants) share the same customer environment and differ in the key facts behind the problem; there are 95 tasks in all. The last column gives the held-out data on which the original task is evaluated.}
\centering\footnotesize\setlength{\tabcolsep}{2.5pt}
\begin{tabular*}{\linewidth}{@{\extracolsep{\fill}}lllrl@{}}
\toprule
\textbf{Scenario} & \textbf{What is delivered} & \textbf{Sector} & \textbf{Tasks} & \textbf{Evaluated on} \\
\midrule
\texttt{bank\_scam\_hold} & Anti-scam hold queue agent & Banking & 3 & 523 queue calls \\
\texttt{build\_pay\_cutover} & Subcontract payment-ledger cut-over & Construction & 4 & one cut-over and its claims \\
\texttt{campus\_booking} & Space-booking assistant & Education & 5 & 82 conversations \\
\texttt{charge\_care\_config} & No-code driver-support agent & EV charging & 4 & the customer’s test suite \\
\texttt{estate\_repair\_cases} & Agent for fund-paid major repairs & Property mgmt. & 4 & 145 cases over 63 days \\
\texttt{fund\_recon} & Reconciliation break clearing & Asset management & 8 & 188 breaks \\
\texttt{grid\_interval\_pipeline} & Nightly interval-settlement job & Power grid & 4 & nightly settlement runs \\
\texttt{hiring\_screen} & Applicant screening tool & Logistics & 3 & two site configurations \\
\texttt{hospital\_triage} & Triage chat with a trial run & Healthcare & 4 & 190 sessions \\
\texttt{listing\_handover} & Listing compliance checker & E-commerce & 4 & 345 listings \\
\texttt{logistics\_claims} & After-sales agent that can pay out & Logistics & 7 & 212 tickets \\
\texttt{mfg\_chatbi} & “Ask the data” assistant & Manufacturing & 7 & 94 live-database questions \\
\texttt{news\_incident} & AI flash desk for a newsroom & Media & 6 & two weeks of stories \\
\texttt{oil\_npt} & Drilling downtime coder & Energy & 5 & 90 drilling reports \\
\texttt{oncall\_incident} & AI night responder & Software & 3 & 184 incidents \\
\texttt{prior\_auth} & Prior-authorization agent & Health insurance & 3 & 165 requests \\
\texttt{revenue\_dispute} & Monthly revenue report & Manufacturing & 3 & 100 disputed items \\
\texttt{review\_screening} & Systematic-review screening & Research & 6 & 1,400 records \\
\texttt{robot\_cell\_rollout} & Mobile-manipulator rollout plan & Warehouse robotics & 5 & 12 simulated months \\
\texttt{water\_twin} & Reservoir release rule & Water & 7 & six scenario years \\
\bottomrule
\end{tabular*}
\end{table}

\subsection{Admission Anchors and Anti-Gaming Validation}

Each task included in FDE-Gym must satisfy four predefined scoring anchors (Figure 3). (1) The reference solution must score at least 0.95. (2) The do-nothing baseline must score 0. (3) The literal solution, which implements exactly what the task instruction asks for without seeking any further constraint, must score at most 0.2. This condition holds even when the sponsor’s request identifies the right problem, because the constraints omitted from the task instruction still determine the outcome. The only exception is a task in which the sponsor’s request is both correct and complete; there the literal solution coincides with the reference solution and must score at least 0.95, so that distrusting the sponsor is never a universally rewarded strategy. (4) Shortcut strategies must score at most 0.8. These strategies include overfitting to historical data, guessing round-number thresholds, and deferring all uncertain decisions to humans. Scores are capped near zero if an agent attempts to bypass customer controls or manipulate the grader. The graders were tested against more than 270 shortcut and bypass strategies. Agents that solve tasks using only customer documents and data are scored on the same basis as those that interact with stakeholders.

\subsection{Expert Review and the Online Platform}

Industry experts cross-reviewed and iteratively refined each task. Their total contribution to the development of FDE-Gym exceeded 10,000 hours. FDE-Gym also provides an online platform (\href{https://fde-gym.com}{fde-gym.com}) where human participants use the same workspaces, interact with the same stakeholders, and are evaluated by the same graders as agents. A human baseline is currently being collected there under recorded conditions from industry experts working on the L3 tasks. The results will be reported in a future version of this paper.

\section{Experimental Setup}

\textbf{Models.} We evaluate 13 model configurations from seven providers (12 models, one evaluated both with and without extended thinking): Claude Opus 5.5, Claude Sonnet 5.5, GPT-6 Astra, GPT-6.1 Sol, GLM-5.3 Flash, DeepSeek V4.1 Flash, Gemini 3.8 Flash, Qwen 3.8 Flash, GPT-5.5, GPT-6 Luna, Mistral Small 4, Claude Haiku 4.5, and Claude Haiku 4.5 (thinking). All models use the same minimal agent framework with five tools (execute commands, read files, write files, edit files, and list files), without model-specific customization. They contact stakeholders by invoking a command in the workspace.

\textbf{Protocol.} We evaluate models at four task instruction levels (L0–L3), from a fully specified problem to the sponsor’s original request. L3 is the primary setting, reflecting real-world delivery where the request may not address the underlying problem. It covers all 95 tasks: 20 original tasks and 75 variants with altered key facts. To measure the effect of information provided in the task instruction on performance, we also evaluate L0–L2 on the original tasks. Three scenarios have identical L2 and L3 task instructions; for these, the L3 run on the original task also serves as the L2 result, so all four levels are reported on the same 20 original tasks. Each model is therefore run on 20 tasks at L0, 20 at L1, 17 at L2, and 95 at L3, totaling 152 task–instruction-level combinations per model and 1,976 across all 13 model configurations. Each combination is run three times, and the scores are averaged. Each run ends when the agent submits its delivery or reaches a limit of 200 turns, 150 minutes, or USD 8 in model usage cost, whichever comes first. The system is then deployed and evaluated in the state reached at that point.

\textbf{Reporting.} We report the mean FDE-Gym score across evaluated tasks at each task instruction level, with 95\% confidence intervals computed by bootstrapping over tasks for the full L3 evaluation. We also report the normalized business uplift before stakeholder contact cost, to distinguish system performance from stakeholder contact cost, along with the model usage cost per run, as recommended in work on cost-aware evaluation {\color{navy}[\hyperlink{ref.15}{15}, \hyperlink{ref.16}{16}]}. Scores, confidence intervals, normalized business uplift, and stakeholder contact cost are shown in Table 3; mean model usage cost per run is shown in Figure 5b.

\textbf{Comparison with other benchmarks.} We compare FDE-Gym scores with task success rates on Terminal-Bench 2.1 {\color{navy}[\hyperlink{ref.2}{2}]}, Vibe Code Bench {\color{navy}[\hyperlink{ref.12}{12}]}, and Finance Agent {\color{navy}[\hyperlink{ref.13}{13}]}, using results from Vals AI (vals.ai/benchmarks, accessed 6 October 2026). The published results are task success rates, whereas the FDE-Gym result is a normalized uplift; the figure therefore places the two side by side without equating them. We use a single independent source so that each benchmark is run in the same way for every model. The other benchmarks cited in this paper are not run by this source, and SWE-bench Verified is no longer run on new models there, so comparable results are not available for the models shown. For each provider, we show its best-performing model on FDE-Gym among those evaluated in our study with results available on all three external benchmarks. To compare with success rates reported as percentages, we multiply FDE-Gym scores by 100.

\textbf{Availability.} Both the harness that creates task workspaces, simulates stakeholders, and grades deliveries, and the experiment runner used in this study are open source (\href{https://github.com/deeplethe/fde-gym}{github.com/deeplethe/fde-gym}, version 1.1.0). The 20 original tasks and their graders are available on request (\href{https://huggingface.co/datasets/DeepLethe/FDE-Gym}{huggingface.co/datasets/DeepLethe/FDE-Gym}), with three tasks public as examples. The 75 variants and all reference solutions are held out.

\section{Results}

\subsection{Even the Best Model Is Far from the Reference}

\begin{finding}{Finding 1}
The best configuration achieves an average FDE-Gym score of 0.56 and resolves 14 of 95 tasks. Even selecting the best configuration for each task does not close the gap.
\end{finding}

Figure 2 and Table 3 present the main results. At L3, Claude Opus 5.5 leads with 0.56 (95\% interval 0.50 to 0.62), followed by Claude Sonnet 5.5 at 0.49, GPT-6 Astra at 0.38, and GPT-6.1 Sol at 0.37. We refer to the first three as the three leading models throughout, with GPT-6.1 Sol just below them. Three further configurations score near 0.2, two are only marginally above zero, and the remaining four score below zero. No model resolves more than 15 of 95 tasks, and five of the thirteen configurations resolve none.

The shortfall is not driven by a few unusually difficult scenarios. On only one scenario, \texttt{bank\_scam\_hold}, does any model average at least 0.8 (Figure 7). Taking, for each task, the best score achieved by any of the thirteen configurations yields 0.70 with 30 tasks resolved; on 16 tasks no configuration reaches 0.5.

The upper end of the scale is attainable: every reference solution scores at least 0.95 under the same grader, and when the rules are provided (L0) the leading models reach 0.73–0.82. How far human experts can proceed when given only the sponsor’s request will be established by the human baseline currently being collected.

\begin{table}[tbp]
\caption{Main results. Left: score by task instruction level on the 20 original tasks. Right: all 95 tasks at L3, with the 95\% interval, the normalized business uplift before stakeholder contact cost, the stakeholder contact cost, and the number of tasks resolved (score \ensuremath{\geq} 0.8, no incident) and the number of tasks finished below the do-nothing baseline (score \ensuremath{<} 0).}
\centering\small\setlength{\tabcolsep}{2.5pt}
\begin{tabular*}{\linewidth}{@{\extracolsep{\fill}}lrrrrrrrrrr@{}}
\toprule
 & \multicolumn{4}{c}{\textbf{Original tasks}} & \multicolumn{6}{c}{\textbf{All 95 tasks, L3}} \\
\cmidrule(lr){2-5} \cmidrule(lr){6-11}
\textbf{Configuration} & \textbf{L0} & \textbf{L1} & \textbf{L2} & \textbf{L3} & \textbf{Score} & \textbf{95\% CI} & \textbf{U} & \textbf{C} & \textbf{Resolved} & \textbf{\ensuremath{<} 0} \\
\midrule
Claude Opus 5.5 & 0.82 & 0.62 & 0.54 & 0.55 & \textbf{0.56} & 0.50, 0.62 & 0.70 & 0.14 & 14 & 5 \\
Claude Sonnet 5.5 & 0.74 & 0.45 & 0.47 & 0.46 & \textbf{0.49} & 0.43, 0.55 & 0.67 & 0.18 & 15 & 7 \\
GPT-6 Astra & 0.73 & 0.39 & 0.34 & 0.33 & \textbf{0.38} & 0.30, 0.46 & 0.60 & 0.22 & 9 & 18 \\
GPT-6.1 Sol & 0.76 & 0.33 & 0.28 & 0.43 & \textbf{0.37} & 0.28, 0.44 & 0.57 & 0.20 & 7 & 17 \\
GLM-5.3 Flash & 0.68 & 0.29 & 0.38 & 0.24 & \textbf{0.22} & 0.11, 0.31 & 0.40 & 0.18 & 8 & 28 \\
DeepSeek V4.1 Flash & 0.61 & 0.20 & 0.27 & 0.28 & \textbf{0.21} & 0.10, 0.30 & 0.38 & 0.18 & 9 & 26 \\
Gemini 3.8 Flash & 0.58 & 0.25 & 0.15 & 0.21 & \textbf{0.19} & 0.09, 0.28 & 0.45 & 0.26 & 4 & 28 \\
Qwen 3.8 Flash & 0.36 & 0.06 & 0.09 & 0.07 & \textbf{0.07} & \ensuremath{-}0.03, 0.17 & 0.27 & 0.21 & 4 & 39 \\
GPT-5.5 & 0.51 & 0.15 & 0.04 & 0.06 & \textbf{0.03} & \ensuremath{-}0.05, 0.11 & 0.11 & 0.08 & 0 & 44 \\
GPT-6 Luna & 0.12 & \ensuremath{-}0.02 & \ensuremath{-}0.07 & \ensuremath{-}0.12 & \textbf{\ensuremath{-}0.12} & \ensuremath{-}0.20, \ensuremath{-}0.05 & 0.00 & 0.12 & 0 & 63 \\
Mistral Small 4 & \ensuremath{-}0.07 & \ensuremath{-}0.25 & \ensuremath{-}0.31 & \ensuremath{-}0.31 & \textbf{\ensuremath{-}0.32} & \ensuremath{-}0.41, \ensuremath{-}0.23 & \ensuremath{-}0.31 & 0.01 & 0 & 57 \\
Claude Haiku 4.5 & 0.11 & \ensuremath{-}0.11 & \ensuremath{-}0.30 & \ensuremath{-}0.25 & \textbf{\ensuremath{-}0.33} & \ensuremath{-}0.42, \ensuremath{-}0.24 & \ensuremath{-}0.23 & 0.09 & 0 & 79 \\
Claude Haiku 4.5 (thinking) & 0.14 & \ensuremath{-}0.23 & \ensuremath{-}0.29 & \ensuremath{-}0.24 & \textbf{\ensuremath{-}0.35} & \ensuremath{-}0.44, \ensuremath{-}0.26 & \ensuremath{-}0.27 & 0.08 & 0 & 74 \\
\bottomrule
\end{tabular*}

\smallskip\raggedright\footnotesize U: normalized business uplift; C: stakeholder contact cost. For three scenarios whose L2 and L3 task instructions are identical, the L3 result is also reported as the L2 result.
\end{table}

\subsection{Withholding Constraints Produces the Largest Observed Score Drop}

\begin{finding}{Finding 2}
Withholding the constraints reduces the scores of the three leading models by 0.28 on the same tasks (95\% interval 0.17 to 0.39). Making the request vague, and then misdirected, reduces them by only a further 0.04 (\ensuremath{-}0.03 to 0.14).
\end{finding}

Figure 2b traces the progression across task instruction levels, and the left half of Table 3 reports every level on the same 20 original tasks; the values quoted in this section refer to these tasks. With the rules given (L0), the three leading models score 0.73–0.82 and resolve 11–13 of 20 tasks. At L1, where the problem is still stated correctly but its constraints must be found, they score 0.39–0.62 and resolve 1–7. The loss has two components. The delivered systems deteriorate, from 0.81–0.90 to 0.63–0.80 of the reference uplift, and discovering the constraints incurs a cost: questions per task rise from 3.9–7.0 to 11.3–13.0, and stakeholder contact cost from 0.07–0.10 to 0.18–0.24.

Beyond this step, the scores of these models are nearly flat. Claude Opus 5.5 scores 0.62, 0.54, and 0.55 at L1, L2, and L3; Claude Sonnet 5.5 scores 0.45, 0.47, and 0.46; and GPT-6 Astra scores 0.39, 0.34, and 0.33. Using paired results for the same 20 tasks and averaging across the three models, the decline from L0 to L1 is 0.28 (95\% interval 0.17 to 0.39) and the decline from L1 to L3 is 0.04 (\ensuremath{-}0.03 to 0.14). The first step therefore costs more than the two that follow, although a further loss of up to 0.14 cannot be excluded, and for an individual model the upper bound is as large as 0.25. Our initial hypothesis was that problem definition would be the bottleneck: an agent given a symptom or a wrong solution would build the wrong system. For the leading models, the data do not support this. One interpretation is that once these models must search for constraints at all, they also discover what the sponsor actually needs. For weaker models, the steps beyond L1 remain costly: across L0–L3, GPT-5.5 scores 0.51, 0.15, 0.04, and 0.06, and Claude Haiku 4.5 scores 0.11, \ensuremath{-}0.11, \ensuremath{-}0.30, and \ensuremath{-}0.25. %

The fully specified level yields a result of its own: even with every rule provided, no model exceeds 0.82 on the same 20 tasks. Knowing the facts does not remove the need for engineering.

\subsection{Recovering Key Facts Does Not Ensure Effective Delivery}

\begin{finding}{Finding 3}
Leading models obtain most of the facts held by stakeholders yet deliver only 0.60–0.70 of the reference uplift; stakeholder contact costs further reduce their scores.
\end{finding}

The grader records which key facts an agent obtained from stakeholders, which allows us to separate information discovery from system building. At L3 the three leading models ask 10.0–12.3 questions per task and obtain 72–85\% of the key facts. Their normalized business uplift is 0.60–0.70. Figure 5a shows the pattern across all thirteen configurations: more facts are associated with better systems, but the leading group sits at the far right of the plot and still well below the reference solution. In the 73 task–model cases where one of the three obtained \textit{every} key fact, the normalized business uplift averages 0.66 and reaches at least 0.8 in 27 of them (37\%). For Claude Opus 5.5 alone, every key fact was obtained on 23 tasks, and the uplift reached at least 0.8 on 10 of them.

\begin{figure}[tbp]
\centering
\includegraphics[width=1.000\linewidth]{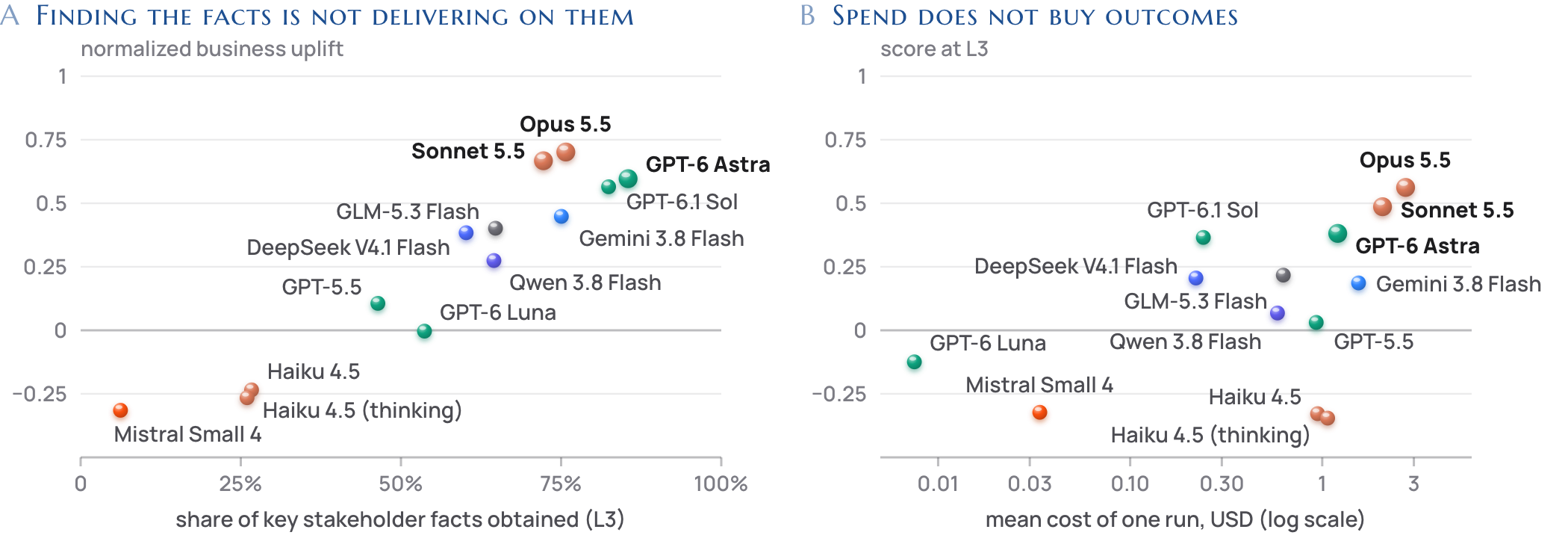}
\caption{\textbf{(a)} Share of key stakeholder facts obtained against normalized business uplift, at L3. Leading models learn most of the facts and deliver 0.60–0.70 of the reference uplift. \textbf{(b)} Mean model usage cost per run against score at L3.}
\end{figure}

The second component of the finding is stakeholder contact cost. GPT-6 Astra’s delivered systems reach 0.8 on 36 of 95 tasks, but only 9 remain resolved once stakeholder contact cost is deducted; for Claude Opus 5.5 the figures are 40 and 14. Agents are not informed that questions incur a cost, and they ask as if questions were free. Stakeholder contact cost consumes 0.14 of the 0.70 uplift delivered by Claude Opus 5.5 and 0.22 of the 0.60 delivered by GPT-6 Astra. At the other extreme are models that rarely ask: Mistral Small 4 asks 0.7 questions per task and obtains 6\% of the key facts.

\subsection{Weak Models Frequently Score Below the Do-Nothing Baseline}

\begin{finding}{Finding 4}
The weakest models score below the do-nothing baseline on 60–83\% of tasks; on these tasks, delivering nothing would have been the better outcome.
\end{finding}

\begin{figure}[tbp]
\centering
\includegraphics[width=0.843\linewidth]{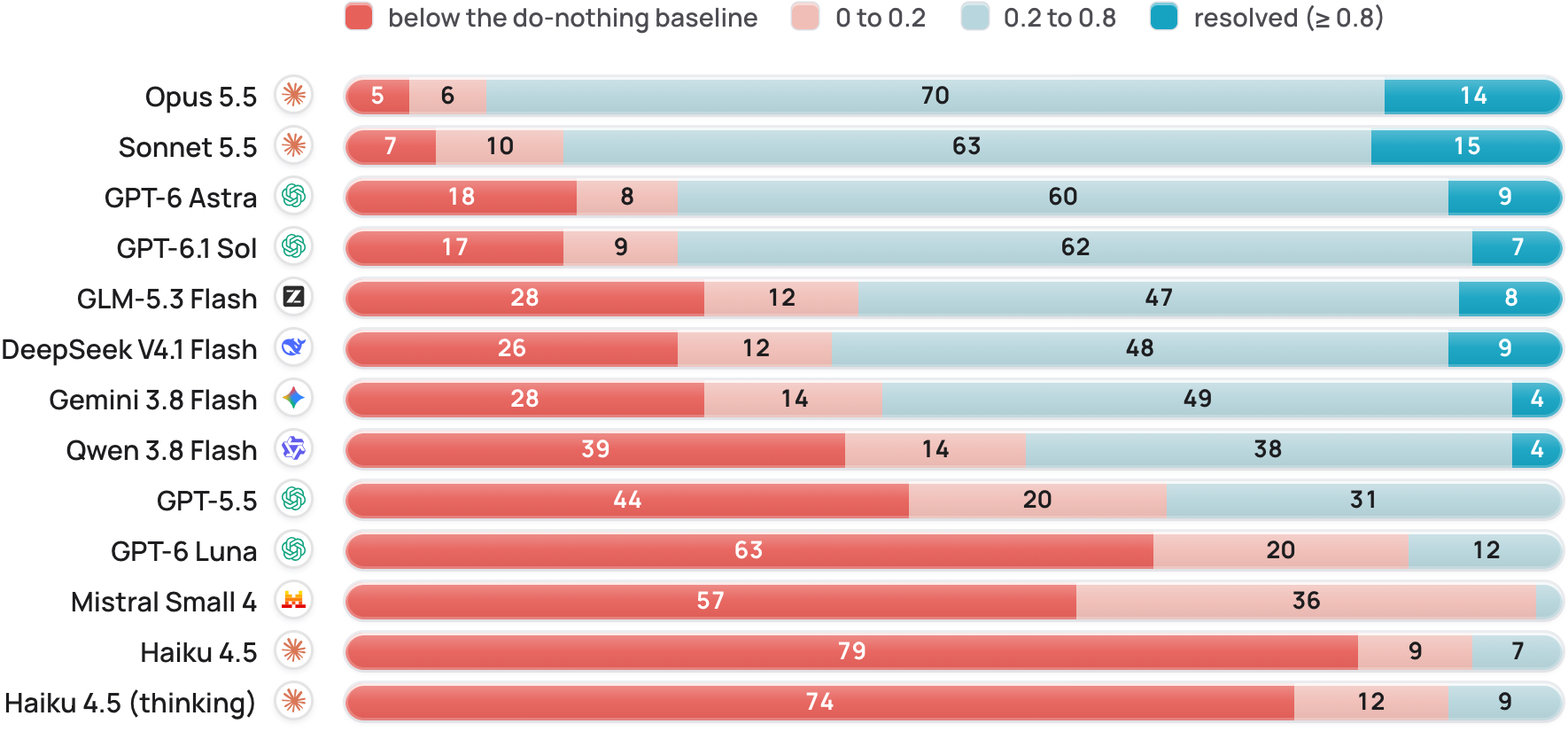}
\caption{Distribution of each model’s 95 tasks at L3 by score band. Scores of 0.2 or below correspond to the cap on literal compliance with the sponsor’s task instruction.}
\end{figure}

A score of zero is not the floor: a system that pays out claims it should have refused, charges the wrong party, or wakes the wrong engineer performs worse than the system it replaced. Figure 6 sorts each model’s 95 tasks at L3 into four score bands. Claude Opus 5.5 scores below zero on 5 tasks and Claude Sonnet 5.5 on 7, whereas Claude Haiku 4.5 does so on 79 and GPT-6 Luna on 63.%

Part of this reflects stakeholder contact cost, but much of it reflects the delivered system itself. Before any stakeholder contact cost is deducted, the systems delivered by Mistral Small 4 score below the do-nothing baseline on 51 tasks and those delivered by Claude Haiku 4.5 (thinking) on 55, compared with 1–4 for the three leading models. On 23 tasks, the system delivered by Mistral Small 4 caused more harm than the benefit produced by the reference solution. Extended thinking does not help: Claude Haiku 4.5 scores \ensuremath{-}0.33 without it and \ensuremath{-}0.35 with it. For a buyer, the practical implication is that a weak agent is not a cheaper version of a strong one; its expected contribution on these tasks is negative. %

\subsection{Model Performance Varies Substantially Across Scenarios and Task Variants}

\begin{finding}{Finding 5}
On five of the 20 scenarios no model averages at least 0.5, and the same scenario with different key facts yields a different score.
\end{finding}

\begin{figure}[tbp]
\centering
\includegraphics[width=0.891\linewidth]{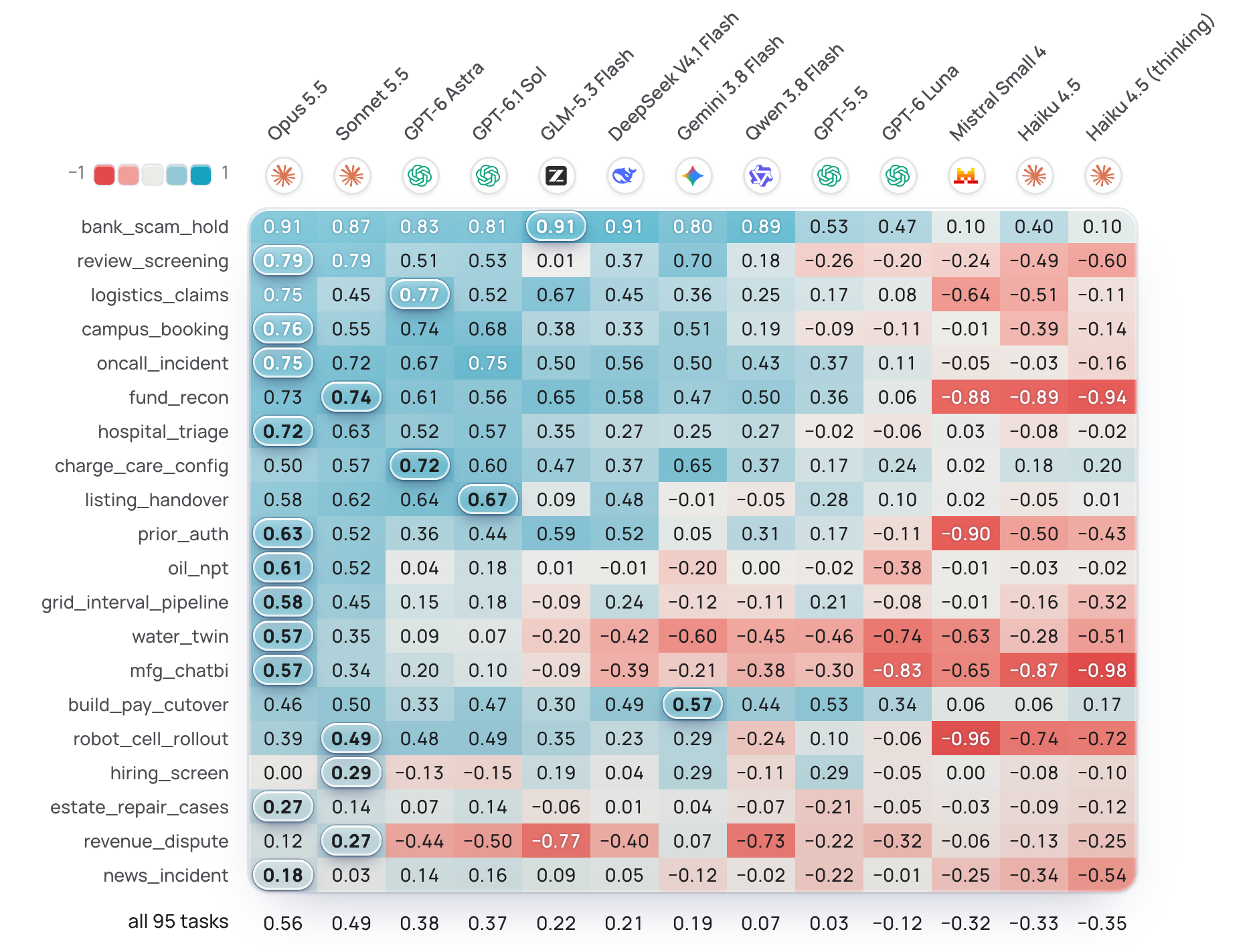}
\caption{Mean score at L3 on each scenario, for each model; scenarios are ordered by the best model’s score, and the highest-scoring cell in each row is outlined. Color runs from red (below the do-nothing baseline) to blue (at the reference solution).}
\end{figure}

Figure 7 breaks the headline result down by scenario. Claude Opus 5.5 is the best-performing model on 11 scenarios, Claude Sonnet 5.5 on four, GPT-6 Astra on two, and GLM-5.3 Flash, Gemini 3.8 Flash, and GPT-6.1 Sol on one each. The spread within a row can be wide even among the leading models: on \texttt{revenue\_dispute} the three score 0.12, 0.27, and \ensuremath{-}0.44. On five scenarios no model averages at least 0.5, and on the two hardest, \texttt{news\_incident} and \texttt{revenue\_dispute}, the best scores are only 0.18 and 0.27.

The variants show how strongly a score depends on the particular facts of a task. Across the tasks within a scenario, the median range of a leading model’s score is 0.34–0.44, and this range exceeds 0.5 in 6–8 of the 20 scenarios.

\subsection{Higher Spending and Effort Do Not Consistently Correspond to Higher Scores}
Cost and effort explain little of the ranking (Figure 5b). GPT-6.1 Sol spends USD 0.24 and 25 turns per run and scores 0.37, whereas Gemini 3.8 Flash spends USD 1.54 and 121 turns and scores 0.19. Qwen 3.8 Flash works longest, at 169 turns and 110 minutes per run, submits within the limits on only 11 of its 152 task–instruction-level combinations, and scores 0.07; GLM-5.3 Flash submits on 95. The three leading models submit on every combination, in 26–60 turns. Recognizing when the work is complete appears to be part of the required skill. %

\subsection{Comparison with other benchmarks}
\begin{figure}
\centering
\includegraphics[width=0.8\linewidth]{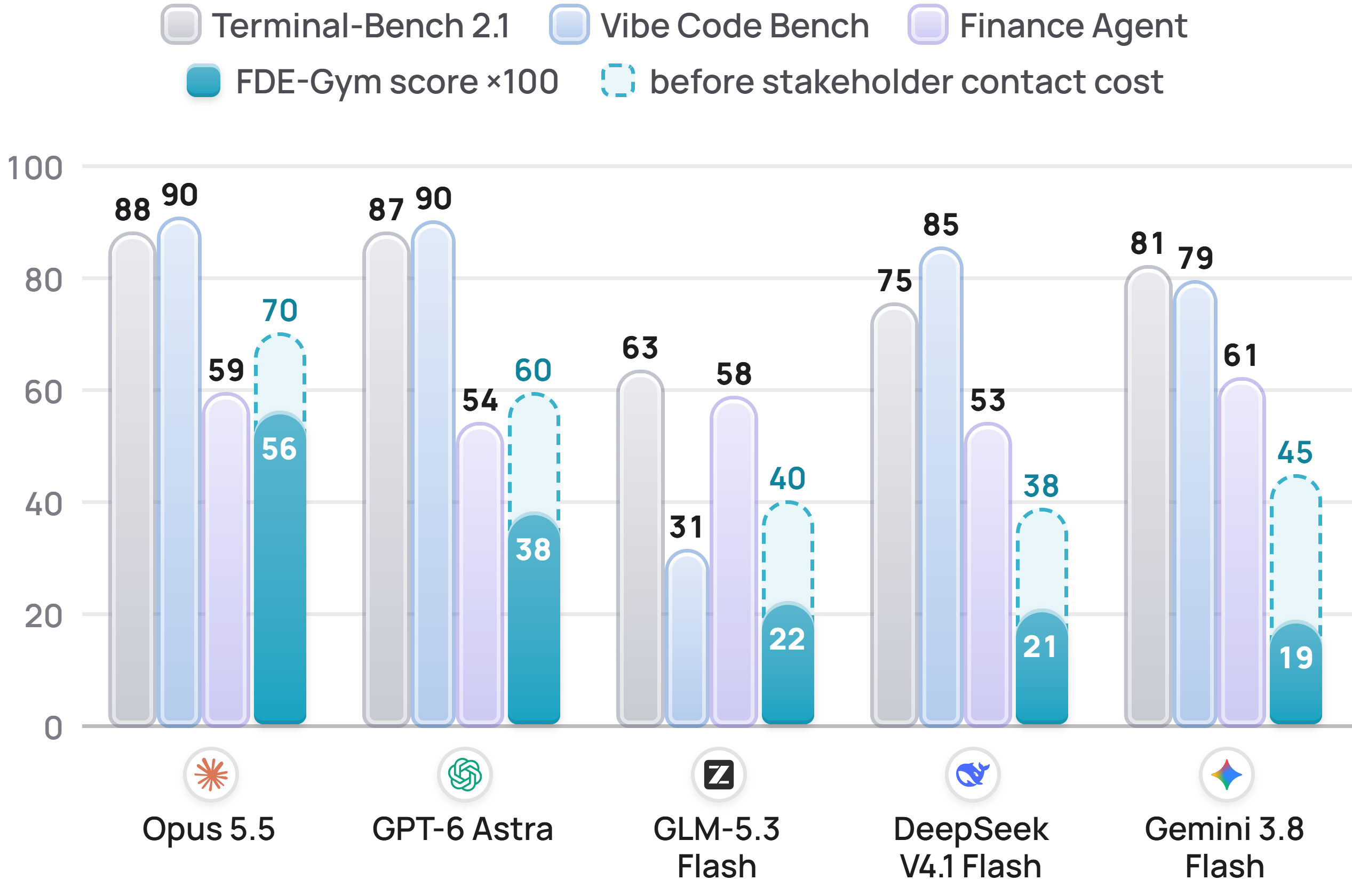}
\caption{Model performance on Terminal-Bench 2.1, Vibe Code Bench, Finance Agent, and FDE-Gym.}
\end{figure}

Figure 8 compares FDE-Gym with three public benchmarks covering work that precedes delivery. These benchmarks evaluate engineering tasks in a terminal (Terminal-Bench 2.1 {\color{navy}[\hyperlink{ref.2}{2}]}), complete web application development from a specification (Vibe Code Bench {\color{navy}[\hyperlink{ref.12}{12}]}), and research tasks performed by financial analysts (Finance Agent {\color{navy}[\hyperlink{ref.13}{13}]}). Dashed outlines show normalized business uplift before deducting stakeholder contact cost. Models that solve up to 88\%, 90\%, and 61\% of tasks on these three benchmarks, respectively, score 0.19–0.56 on FDE-Gym. Because task success rates and normalized business uplift are different quantities, the figure places them side by side on their respective scales without treating the difference as a directly comparable gap. The closest counterpart to a success rate in FDE-Gym is the proportion of resolved tasks (a score of at least 0.8 with no incident). Among the models shown in Figure 8, the leading model Claude Opus 5.5 resolves 14 of 95 tasks (15\%). For a customer, the practical implication is that a model’s performance on component benchmarks is a weak guide to what it will deliver: the same models that solve most terminal, coding, and research tasks improve the customer’s business metric in at most one of seven engagements.

\section{Related Work}

Existing benchmarks each measure one stage of a delivery. Table 4 arranges them by the capability they isolate. To our knowledge, none joins these stages into a single causal chain in which a misreading of the problem renders good code worthless and the loss appears on a business metric.

The most closely related work was published within weeks of ours. \ensuremath{\tau}\textsuperscript{\ensuremath{\tau}}-bench {\color{navy}[\hyperlink{ref.9}{9}]} makes agent construction the task: a developer agent receives a business’s records, a client who holds the requirements, an existing codebase, and limits on serving cost, and the customer-service agent it delivers is then deployed against held-out simulated users. Its strongest configuration passes 23.9\% of simulations, compared with 82.2\% for an expert-authored reference; we likewise find the best model well short of a reference written by experts. FDE-Gym differs in what it varies and in what it measures. The task instruction varies from a full statement of the rules to a request that may be misdirected, stakeholder contact incurs a charge, the result is measured as uplift on the customer’s business metric relative to a do-nothing baseline of zero, which a delivery can fall below, and the scenarios span 18 sectors and deliverables beyond customer-service agents. Trains but Doesn’t Learn {\color{navy}[\hyperlink{ref.10}{10}]} places an agent in the role of a forward-deployed engineer for model post-training and identifies the failure mode in which every check continues to pass while the delivered model is no better than the base model; our literal-compliance anchor is designed to expose the same kind of delivery.

\begin{table}[tbp]
\caption{Prior benchmarks by the delivery capability they measure, and what FDE-Gym adds to each.}
\centering\footnotesize
\begin{tabularx}{\linewidth}{@{}l>{\raggedright\arraybackslash}X>{\raggedright\arraybackslash}X>{\raggedright\arraybackslash}X@{}}
\toprule
\textbf{Capability} & \textbf{Representative benchmarks} & \textbf{What they measure} & \textbf{What FDE-Gym adds} \\
\midrule
Framing & Ambig-SWE {\color{navy}[\hyperlink{ref.7}{7}]}, LHAW {\color{navy}[\hyperlink{ref.8}{8}]}, SWE-RPG {\color{navy}[\hyperlink{ref.17}{17}]}, UserBench {\color{navy}[\hyperlink{ref.18}{18}]}, ReqElicitGym {\color{navy}[\hyperlink{ref.19}{19}]}, XYEval {\color{navy}[\hyperlink{ref.11}{11}]} & Whether agents ask when information is missing, elicit what a user has not said, or resist a misleading suggestion & A misdirected request inside a full engagement; a link from the framing error to a business outcome \\[3pt]
Discovery & TheAgentCompany {\color{navy}[\hyperlink{ref.5}{5}]}, Workspace-Bench {\color{navy}[\hyperlink{ref.20}{20}]}, RUBICON {\color{navy}[\hyperlink{ref.21}{21}]}, DRBench {\color{navy}[\hyperlink{ref.22}{22}]}, Implicit Intelligence {\color{navy}[\hyperlink{ref.23}{23}]} & Finding information and unstated constraints across systems, files, and simulated colleagues & Sources that contradict one another, become outdated, or represent a particular position; a stakeholder contact cost for asking \\[3pt]
Solution design & WorkArena++ {\color{navy}[\hyperlink{ref.24}{24}]}, CRMArena-Pro {\color{navy}[\hyperlink{ref.6}{6}]}, EnterpriseBench {\color{navy}[\hyperlink{ref.25}{25}]}, APEX-Agents {\color{navy}[\hyperlink{ref.26}{26}]} & Completing knowledge work and operations in enterprise systems & Whether the solution fits the customer’s organization and constraints \\[3pt]
Implementation & SWE-bench {\color{navy}[\hyperlink{ref.1}{1}]}, SWE-Bench Pro {\color{navy}[\hyperlink{ref.27}{27}]}, Terminal-Bench {\color{navy}[\hyperlink{ref.2}{2}]}, SWE-Lancer {\color{navy}[\hyperlink{ref.3}{3}]} & Writing code that passes hidden tests & Whether the code was the right thing to write \\[3pt]
Deployment and operations & FDE-Bench {\color{navy}[\hyperlink{ref.4}{4}]}, ITBench {\color{navy}[\hyperlink{ref.28}{28}]}, AIOpsLab {\color{navy}[\hyperlink{ref.29}{29}]} & Building deployable artifacts that pass gated checks; diagnosing and mitigating incidents & Customer interaction, requirements discovery, and business outcomes \\[3pt]
Delivery & \ensuremath{\tau}\textsuperscript{\ensuremath{\tau}}-bench {\color{navy}[\hyperlink{ref.9}{9}]}, Trains but Doesn’t Learn {\color{navy}[\hyperlink{ref.10}{10}]} & Building an agent, or a fine-tuned model, for a client and judging what was delivered after deployment & Task instructions varying from a full statement of the rules to a request that may be misdirected; uplift on a business metric minus stakeholder contact cost; many sectors and forms of deliverable \\[3pt]
Outcome & Vending-Bench {\color{navy}[\hyperlink{ref.30}{30}]}, E-Commerce Bench {\color{navy}[\hyperlink{ref.31}{31}]}, GDPval {\color{navy}[\hyperlink{ref.32}{32}]}, Remote Labor Index {\color{navy}[\hyperlink{ref.33}{33}]} & Profit from running a simulated business; expert judgment of a deliverable & An engineering deliverable that produces the measured outcome; reproducibility without human judges \\[3pt]
\bottomrule
\end{tabularx}
\end{table}

Two lines of work bear directly on our task instruction levels. XYEval {\color{navy}[\hyperlink{ref.11}{11}]} turns existing benchmarks, \ensuremath{\tau}\textsuperscript{2}-bench {\color{navy}[\hyperlink{ref.34}{34}]} among them, into XY problems, in which the user asks about an attempted solution rather than the real problem, and reports relative drops of up to 46.7\%. Our leading models lose little when the sponsor’s request is misdirected, perhaps because, within a full engagement, the documents, data, and stakeholders around the sponsor are all available to contradict it. UnderSpecBench {\color{navy}[\hyperlink{ref.35}{35}]} holds the environment and the correct action fixed while degrading the instruction along three axes, an isolation strategy that our task instruction levels also use, and finds that underspecification leads agents to guess rather than fail; Implicit Intelligence {\color{navy}[\hyperlink{ref.23}{23}]} reports a best pass rate of 48.3\% on simple requests whose constraints must be discovered. Both are consistent with where our levels locate the largest drop: at the constraints that nobody stated.

Three of the benchmarks in Table 4 shaped FDE-Gym directly. TheAgentCompany {\color{navy}[\hyperlink{ref.5}{5}]} is the closest in form, a self-contained simulated company with colleagues to talk to; we keep the self-contained world and replace “a colleague assigned this task” with “a customer hired you”, and task completion with a business metric. LHAW {\color{navy}[\hyperlink{ref.8}{8}]} is the most systematic treatment of missing information and models the cost of asking; our L0 to L1 step reproduces its finding that agents rarely recover full-specification performance by asking, and our L3 adds a situation it does not cover: a request that is misdirected. Vending-Bench {\color{navy}[\hyperlink{ref.30}{30}]} showed that an executable business simulation can replace human judges; we attach one to an engineering deliverable.

Two methodological warnings also shaped the design. Results obtained against LLM-played users depend on which model plays the user {\color{navy}[\hyperlink{ref.14}{14}]}, which is why our stakeholders are scripted, and single-run success overstates reliability {\color{navy}[\hyperlink{ref.36}{36}, \hyperlink{ref.37}{37}]}, which is why every task is run three times. The admission anchors follow the practice of auditing an agentic benchmark with do-nothing and shortcut solutions before trusting its scores {\color{navy}[\hyperlink{ref.38}{38}]}. Beyond single benchmarks, long-horizon evaluations track the length of software tasks agents can complete {\color{navy}[\hyperlink{ref.39}{39}]}, and Collaborative Gym {\color{navy}[\hyperlink{ref.40}{40}]} studies agents working alongside people; FDE-Gym instead sets people and agents the same tasks separately, so that the two can be compared.

\section{Limitations}

\textbf{A simulated customer.} Stakeholders are scripted so that results are reproducible. Real people are less predictable: they change their minds, answer a different question from the one asked, and may respond differently on the basis of previous interactions. The contact costs are set by the experts who wrote each scenario and are a model of the cost of stakeholder time, not a measurement of it.

\textbf{Sample size across levels.} Levels L0 to L2 are reported on the 20 original tasks only, and for three of these the L2 result is taken from the identical L3 task instruction. The drop from L0 to L1 is large relative to this sample; the differences among L1, L2, and L3 are not large, and we claim only that no large further drop is visible.

\textbf{No human baseline yet.} This version reports models only. The scale is anchored by the do-nothing baseline and the reference solution, which shows that each task is solvable with full knowledge of the underlying facts; how far human experts can proceed when given only the sponsor’s request has not yet been measured.

\textbf{A single agent framework.} All models run in the same minimal agent framework. The results describe these models in that framework, not the products built around them, and a stronger framework may change both the scores and the ranking.

\textbf{Isolation and cost.} Runs were executed on several hosts with process-level isolation of varying strength; each run’s record is audited after grading, but one container per run would provide a stronger guarantee. Model usage cost is as metered on the route through which each model was accessed and is not a comparison of list prices.

\textbf{Coverage and contamination.} Twenty scenarios across 18 sectors are a sample of delivery work, weighted toward agents and workflows in regulated operations. The scenarios are adapted from real deliveries, and although companies are fictional and variants change the facts, we cannot rule out that related material is in the training data.

\section{Conclusion}

FDE-Gym measures an agent's delivery the way a customer does: not by whether a system was delivered, but by whether the customer's business improved. Measured in this way, the last mile of AI delivery remains long. The best of the thirteen configurations has an average FDE-Gym score of 0.56 and resolves about one task in seven. The gap is not where we expected it to be. Leading models show little performance loss when the sponsor’s request is misdirected, and they obtain most of the facts held by stakeholders. Their principal performance losses arise when translating those facts into an effective system and from the stakeholder contact costs incurred during delivery. Weaker models lose more: on most tasks, the customer would have been better served by no delivery at all.

These are the capabilities an experienced engineer brings to a customer. Closing the gap is therefore a question of judgment in delivery, and FDE-Gym makes that judgment measurable: one number per task, on the customer’s own metric, for people and agents under the same rules.

\section*{Acknowledgments and Disclosure}

\textbf{Competing interests.} DeepLethe develops and operates the FDE-Gym online platform and maintains its leaderboard. None of the evaluated models is developed by an organization with which any author is affiliated.%

\textbf{Anonymous contributors.} Several industry experts who contributed to the scenarios are not listed as authors because of confidentiality obligations, the sensitivity of the projects involved, their employers’ policies, or their own preferences. We are grateful for their expertise and their contributions to this benchmark.

{\footnotesize\raggedright \textbf{Image credits (Figure 1).} Photographs from Wikimedia Commons, cropped. Energy: “North Sea Oil Rig”, Gary Bembridge, CC BY 2.0; Healthcare: “Medical Care and Compassion in a Hospital Bed provided by Nova Ukraine”, NUSF, CC0; Robotics: “BRS2 Amazon fulfilment centre in Swindon”, Auledas, CC BY 4.0; EV charging: “Flo EV fast charging station at Hingham Depot”, 4300streetcar, CC BY 4.0; Water: “Three Gorges Dam”, Gaynor, CC BY 2.0; Logistics: “UPS worker unloads boxes from semi trailer truck onto conveyor belt”, U.S. Department of Agriculture, Public domain; Construction: “Johor Bahru–Singapore Rapid Transit System construction site”, Renek78, CC0; Power grid: “Switchyard of electric substation 750kV”, Novoklimov, CC BY 4.0; Banking: “Safe Deposit Boxes”, Fixedsun, CC0; Media: “The newsroom at DR”, James Cridland, CC BY 2.0; Manufacturing: “Tesla Motors Assembly Line”, Steve Jurvetson, CC BY 2.0; Property: “Xujiahui Shanghai September 2023”, 4300streetcar, CC BY 4.0.\par}
\end{document}